\documentclass[11pt,a4paper]{article}
\usepackage{coling2018}
\usepackage{times}
\usepackage{url}
\usepackage{latexsym}

\usepackage{color}
\usepackage{tabularx}
\usepackage{multirow}
\usepackage{tikz}
\usepackage{tikz-qtree}

\usepackage[utf8]{inputenc}
\usepackage{xcolor}
\usepackage{pgfplots}
\usetikzlibrary{patterns}
\pgfplotsset{compat=1.14}

\definecolor{bblue}{HTML}{4F81BD}
\definecolor{rred}{HTML}{C0504D}
\definecolor{ggreen}{HTML}{9BBB11}
\definecolor{darkerggreen}{HTML}{9BBB88}
\definecolor{ppurple}{HTML}{9F4C7C}
\definecolor{yyellow}{HTML}{FAF532}

\definecolor{llightgrey}{HTML}{DCDCDC}
\definecolor{ssilver}{HTML}{C0C0C0}
\definecolor{ggray}{HTML}{808080}
\definecolor{bblack}{HTML}{000000}

\newcolumntype{R}{>{\raggedleft\arraybackslash}X}

\title{ISO-Standard Domain-Independent Dialogue Act Tagging for Conversational Agents}

\author{
	Stefano~Mezza, Alessandra~Cervone, Giuliano~Tortoreto, \\ \textbf{Evgeny~A.~Stepanov, Giuseppe~Riccardi} \\
	Signals and Interactive Systems Lab,
	University of Trento, Italy\\
 \texttt{name.surname@unitn.it}
}

\date{}

\begin{document}
\maketitle
\begin{abstract}
Dialogue Act (DA) tagging is crucial for spoken language understanding systems, as it provides a general representation of speakers' intents, not bound to a particular dialogue system. 
Unfortunately, publicly available data sets with DA annotation are all based on different annotation schemes and thus incompatible with each other. Moreover, their schemes often do not cover all aspects necessary for open-domain human-machine interaction.
In this paper, we propose a methodology to map several publicly available corpora to a subset of the ISO standard, in order to create a large task-independent training corpus for DA classification. We show the feasibility of using this corpus to train a domain-independent DA tagger testing it on out-of-domain conversational data, and argue the importance of training on multiple corpora to achieve robustness across different DA categories.
\end{abstract}

\section{Introduction}
\blfootnote{
    %
    %
    %
    %
    %
    
    \hspace{-0.65cm}  
    This work is licensed under a Creative Commons 
    Attribution 4.0 International License.
    License details:
    \url{http://creativecommons.org/licenses/by/4.0/}.
}
The correct interpretation of the intents behind a speaker's utterances plays an essential role in determining the success of a dialogue.
Hence, the module responsible for intents classification lies at the very core of many dialogue systems, both in research  
and industry (e.g. Alexa, Siri).
Moreover, although the task of intent recognition is traditionally linked to task-based systems, recently it has also proved crucial for non task-based conversational systems. According to the results of the Amazon Alexa Prize challenge \cite{ram2017}, the most successful conversational systems in the competition relied on a strong spoken language understanding module, while more than 60\% of the approaches explicitly used intents.

Nevertheless, automatic intent recognition is hard, since participants' intents in a dialogue are implicit. Intent classification has therefore been mostly modeled as a supervised machine learning problem \cite{gupta2006t,xu2013convolutional,yang2016end}, with the consequent definition of intents taxonomies. Over time this led to the creation of expensive annotated resources \cite{price1990evaluation,henderson2014second} with the related time-consuming design of multiple intent schemes. In most cases, however, intents taxonomies are defined specifically for a given application or a dataset and are not generalizable to other systems or tasks, making these resources difficult to reuse (e.g. the popular Air Travel Information Services (ATIS) corpus include heavily domain-dependent intents such as \textit{Airfare} or \textit{Ground Service}). 

Dialogue Acts (DA), also known as speech or communicative acts, represent an attempt to create a formalized and generalized version of intents. As such, DAs have been investigated by the research community for many years \cite{stolcke2006dialogue} and have been applied successfully to many tasks. In particular, their aspiration to generality makes them an appealing option for non task-based application (e.g. more than 20\% of the teams in Amazon Alexa Prize Challenge explicitly used DAs \cite{cervoneroving,bowden2018slugbot}, including the winning team \cite{soundingboard2017}).
Also, in the case of DAs, over the years there have been several efforts to produce publicly available annotated resources \cite{godfrey1992switchboard,amicorpus,verbmobil2} to train DA taggers.
The DA taxonomies created for these resources, albeit arguably more general compared to corpora like ATIS (for example utilizing categories such as \textit{wh-questions}), are still dataset specific; since many of these schemes lack coverage of some crucial aspects of dialogic interaction. Furthermore, given that all these datasets utilize different schemes, these resources are hardly compatible. 

The ISO 24617-2 \cite{buntISOstandard,bunt2012iso}, 
the international ISO standard for DA annotation, represents the first attempt to create a truly domain and task independent scheme. Given its holistic approach compared to previous schemes, ISO 24617-2 can be used as a lingua franca for cross-corpora DA mapping, as confirmed by successful attempts to remap single corpora to the standard \cite{chowdhury2016transfer,fang2012annotation}.

However, there is no reference training set for the standard, since the only public resource currently available with ISO 24617-2 annotation (DialogBank, \cite{bunt2016dialogbank}) is too small to be used to train classifiers. Therefore, most DA tagging research still focuses on in-domain studies on large datasets with incompatible DA annotations \cite{stolcke2006dialogue,ji2016latent}. Moreover, most publicly available corpora are imbalanced with respect to the distribution of various DA dimensions such as \textit{Information Transfer} (e.g. ``What's your favourite book?'') or \textit{Action Discussion} (e.g. ``Tell me the news.''), which are required for successful open-domain conversational systems. 

In this work, we show how to reuse and combine publicly available annotated resources to create a large training corpus for domain-independent DA tagging experiments. We map different corpora using an ISO standard compliant DA taxonomy, following the previous research on the topic \cite{fang2012annotation,petukhova2014interoperability} and we share this resource with the research community.\footnote{The suite of scripts we wrote to map and combine publicly available corpora can be found at \url{https://github.com/ColingPaper2018/DialogueAct-Tagger}} 

In order to investigate the soundness of our approach compared to in-domain models we further experiment with domain-independent DA tagging.
As previously done in the literature we cast the Dialogue Act tagging task as a supervised multi-class classification problem using Support Vector Machines. The correctness of the approach is first tested on the de facto DA tagging standard -- the Switchboard (SWDA) corpus \cite{godfrey1992switchboard}, using the reference training and test sets and achieving performance comparable to the state-of-the-art approaches. Secondly, we experiment with domain-independent DA tagging following the same approach and using our combined resource as a training corpus. The DialogBank corpus, that represents a reference manual DA annotation for the ISO standard, is used for the evaluation of the tagger. To the best of our knowledge this is the first attempt to test automatic DA annotation on this corpus.

The domain-independence and suitability of the tagger  for conversational systems trained on multiple resources is additionally evaluated on two other corpora annotated following our optimized taxonomy (human-machine conversations from the Amazon Alexa Prize Challenge). 
The performances achieved on these three datasets suggest that the training on multiple corpora represents a step forward for DA tagging of open-domain non task-based human-machine conversations.   Finally, we present experiments to investigate the contribution of the different corpora to the performance of the classifiers. The results of our experiments show the importance of utilizing multiple resources to achieve a sound performance across different types of DA categories.
The multi-domain DA tagger presented here was successfully employed in Roving Mind, our open-domain conversational system for the Alexa Prize \cite{cervoneroving}.

\section{State of the Art}
\label{sec:soa}

\subsection{Dialogue Act Annotation Schemes}
\label{subsec:soa-da-schemes} 
The notion of Dialogue Acts can be traced back to the one of illocutionary acts introduced by \cite{austin1962things}. The illocutionary act represents a level of description of an utterance's meaning that goes beyond the purely semantic level (``Is the window open?'') to encompass the intent of the speaker in producing that utterance (``Please, close the window.'').

One of the first DA taxonomies was the one created for the task-based corpus MapTask \cite{anderson1991hcrc} in the early nineties. The MapTask scheme distinguishes between \textit{initiating moves} -- such as giving instructions, explaining, checking information or asking questions -- and \textit{response moves} -- for example acknowledging instructions, answering questions and clarifying information. The corpus also makes a distinction regarding the grammatical and semantic structure of the interactions, classifying, for example \textit{wh-questions}, \textit{yn-questions} and \textit{positive/negative answers}. Although pioneering at the time, the MapTask annotation scheme is very specific to the described scenario, and some of its DAs (e.g. \textit{instruct}, \textit{clarify}, \textit{check}) do not scale well to generic, non task-based conversations. Moreover, its taxonomy was not designed to capture all human behaviours during conversations, and, as a consequence, its coverage for labelling a non task-based interaction is inadequate.

The first attempt to define a unified, non task-based standard for DA tagging was the Discourse Annotation and Markup System of Labeling (DAMSL) \cite{DAMSL} tag-set for the SWDA \cite{godfrey1992switchboard} corpus. 
This annotation scheme proposes a taxonomy of 42 tags, describing both semantic aspects of conversation (\textit{opinion}, \textit{non-opinion}, \textit{preference}, etc.), syntactic aspects (\textit{yn-questions}, \textit{wh-questions}, \textit{declarative questions}, etc.) and behaviours related to the dialogue (\textit{conventional closing}, \textit{hedge}, \textit{backchanneling}, etc.). Nevertheless, the taxonomy still has some issues: tags are mutually exclusive (making it impossible to annotate, for example, a no answer which was also signaling non-understanding) and are organised in a flat taxonomy, which does not take into account similarities and differences between the tags.

\newcite{bunt1999dynamic} introduced the Dynamic Interpretation Theory (DIT) for dialogues, setting the theoretical foundation for a domain-independent and task-independent DA taxonomy. The paper introduced some very important concepts like the idea of \textit{multidimensionality} of DAs and the distinction between \textit{Action-Discussion}, a macro-category of DAs encompassing cases in which interlocutors negotiate actions to be performed (e.g. requests like ``Let's switch topic.''), and \textit{Information-Transfer} interactions, capturing the DAs through which speakers exchange information (e.g. sharing personal information like ``My name is John.''). The DIT++ taxonomy \cite{bunt2009dit++} was then defined in 2009 with the aim of providing a unique and universally recognized standard for DA annotation based on the theoretical ideas introduced in the DIT scheme. Its fifth version was accepted as ISO 24617-2. 

The core aspects of the ISO standard are its multidimensionality and its domain and task independence. The ISO scheme is multidimensional in the sense that it makes a clear distinction between \textit{semantic dimensions} (i.e. the aspect of the communication which the DA describes) and \textit{communicative functions} (i.e. the illocutionary act performed within that dimension). In this way, ambiguities between various aspects of the communication and overlapping between DAs are removed. Furthermore, the scheme contains a generic dimension and communicative functions, which is suitable for mapping virtually any kind of conversation, both task-based and non task-based. Moreover, its multidimensional aspect and hierarchical taxonomy make it extensible and potentially adaptable to specific conversational sets. 

\subsection{Dialogue Act Tagging}
\label{subsec:soa-da-tagging}
The automatic recognition of Dialogue Acts has been addressed by the literature using various machine learning techniques. In particular DA classification has been modeled both as a sequence labeling problem, using techniques such as HMM \cite{stolcke2006dialogue}, neural networks \cite{ji2016latent,kumar2017dialogue} or CRF \cite{quarteroni2011simultaneous}, and as a multi-class classification problem, using for example SVM \cite{QuarteroniR10}. Mentioning and comparing all DA classification approaches is difficult because of the differences in annotation schemes and datasets used. All approaches, however, are usually tested on in-domain data.

One of the most popular datasets for benchmarking is SWDA \cite{godfrey1992switchboard}, a dataset of human-human open-domain telephone conversations. The state-of-the-art on SWDA (77.0\% accuracy on 42 DA tags) was achieved  by \cite{ji2016latent} using deep neural networks. 

The ISO standard \cite{buntISOstandard} can be seen as a generalization of all these annotation schemes. However, there is no available training data for the ISO standard.

\section{Data Sets}
\subsection{Training sets}
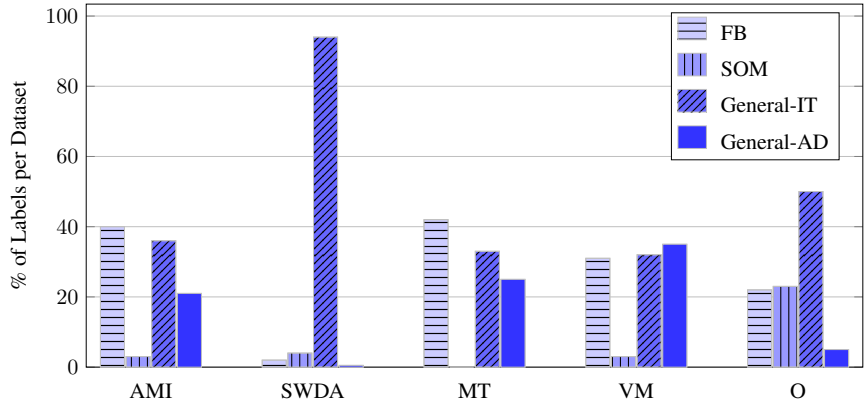
\begin{figure}
\centering
\scalebox{0.75}{
\begin{tikzpicture}
    \begin{axis}[
        width  = 0.95*\textwidth,
        height = 8cm,
        major x tick style = transparent,
        ybar=2*\pgflinewidth,
        bar width=12pt,
        ymajorgrids = true,
        ylabel = {\% of Labels per Dataset},
        symbolic x coords={AMI,SWDA,MT,VM,O},
        xtick = data,
        scaled y ticks = false,
        enlarge x limits=0.10,
        ymin=0,
        enlarge y limits=upper,
        legend cell align=left,
        legend image code/.code={%
                    \draw[#1, draw=none] (0cm,-0.1cm) rectangle (0.5cm,0.3cm);
                }, 
        legend style={
                at={(0.97,0.57)},
                anchor=south east,
                column sep=1ex,
                row sep=0.6ex
        }
    ]
        \addplot +[ssilver,fill=blue!20,mark=none,postaction={
        pattern=horizontal lines
    }]
            coordinates {(AMI, 40.0) (SWDA,2.0) (MT,42.0) (VM, 31.0) (O,22.0)};
        \addplot +[ssilver,fill=blue!40,mark=none,postaction={
        pattern=vertical lines
    }]
             coordinates {(AMI,3.0) (SWDA,4.0) (MT,0.0) (VM, 3.0) (O,23.0)};

        \addplot +[ssilver,fill=blue!60,mark=none,postaction={
        pattern=north east lines
    }]
             coordinates {(AMI, 36.0) (SWDA,94.0) (MT,33.0) (VM, 32.0) (O,50.0)};

        \addplot +[ssilver,fill=blue!80,,postaction={
    }]
             coordinates {(AMI,21.0) (SWDA,0.5) (MT,25.0) (VM, 35.0) (O,5.0)};

        \legend{FB,SOM,General-IT,General-AD}
    \end{axis}
\end{tikzpicture}}
\label{fig:histograms_functions_dataset}
\caption{The distribution of dialogue act categories (after mapping to ISO standard) in various corpora -- AMI, SWDA, MapTask (MT), VerbMobil (VM) and BT Oasis (O). The represented DA categories are Social Obligations Management (SOM) and Feedback dimension DAs, as well as \textit{Action-Discussion} (AD) and \textit{Information-Transfer} (IT) DAs from general dimension.}
\end{figure}

The scarcity of resources of adequate size annotated with the ISO standard makes it difficult to train a DA tagger for this taxonomy. To the best of our knowledge, the DBOX corpus \cite{petukhova2014dbox} -- the only resource manually annotated using the ISO standard -- is not yet publicly available .
The best possible approach given the current availability of data is to map existing corpora's DA schemes to the ISO scheme. Given the limited -- and often domain-dependent -- annotation scheme of these resources, it is impossible to map enough data to train a DA tagger for the full ISO taxonomy, since some of the ISO dialogue acts have no correspondence in any of the considered corpora. Therefore, we opted for a reduced version of the taxonomy, limiting our research to subsets of the General (Task), Social Obligation Management and Feedback dimensions. So far, we mapped the following five different corpora to our scheme:

\textbf{SWDA}: 
The Switchboard corpus \cite{godfrey1992switchboard} is a dataset of transcribed open-domain telephone conversations. The Switchboard Dialogue Act Corpus (SWDA) is a subset of the Switchboard corpus annotated with DAs. SWDA represents a logical choice when building a training set for a domain-independent DA tagger, as it is a large collection of open-domain, non task-based conversations, and therefore provides a natural similarity to the conversational domain of social bots. Moreover, there are already examples in literature of mappings from the Switchboard corpus to the ISO standard \cite{fang2012annotation}. As visible from Figure \ref{fig:histograms_functions_dataset}, drawbacks of the corpus with respect to the task include its unbalancedness (60\% of utterances are \textit{Information-providing}) and lack of \textit{Action-Discussion} interactions (less than 1\% of overall corpus).

\textbf{AMI}: This corpus contains transcriptions from 100 hours of meeting recordings of the European-funded AMI project (FP6-506811), a consortium dedicated to the research and development of technology \cite{amicorpus}. This dataset presents a reasonably balanced collection of utterances and a taxonomy which shares some similarities with the ISO standard (e.g. distinction between \textit{Action-discussion} and \textit{Information-transfer}). Drawbacks of the corpus include the fact that there are multiple speakers (it is therefore more difficult to capture contextual information) and sometimes its scheme does not map to the leaves of the ISO tree.

\textbf{MapTask}: This is a task-based dialogue corpus collected by the HCRC at the University of Edinburgh \cite{anderson1991hcrc}. Dialogues involve two participants, one with an empty map and one with a route-marked map which must instruct the other speaker to draw the same route. The corpus was chosen due to its abundance of \textit{Action-discussion} interactions (more than 30\% of the overall corpus), which are often lacking in other corpora.

\textbf{VerbMobil}: This is a collection of task-based dialogues released in 1997 \cite{burger2000verbmobil}. A subset of these dialogues is annotated with DAs \cite{verbmobil2}. The scenario involves two speakers, which play respectively the roles of a travel agent and of a client. The client usually provides a set of constraints and requests to be satisfied, while the traveling agent has to ask questions and provide information in order to satisfy the client's requests. Interactions happening within the VerbMobil 2 corpus closely resemble those usually seen with personal assistants, with a user looking for the fulfillment of a task and a serving agent interacting with the user to solve his/her issues making it an appealing addition to our training set.

\textbf{BT Oasis}: The BT Oasis corpus is a collection of task-based conversations involving personal assistance for clients of the British Telecom services \cite{leech2003generic}. The conversations are human-to-human, and usually involve a user who has a problem to solve and an assistant who helps the user solving his issues. The BT Oasis corpus was chosen as part of the training set for its interesting scheme, called SPAAC (Speech Act Annotation scheme for Corpora), which is easily mappable to the ISO standard due to its clear separation of grammatical and illocutionary act.

\subsection{Test sets}
\textbf{DialogBank (DB)}: the DialogBank \cite{bunt2016dialogbank} is a corpus\footnote{\url{https://dialogbank.uvt.nl/}} annotated with ISO 24617-2 which currently contains 15 English dialogues: 3 from MapTask and 3 from TRAINS \cite{traum1996conversational} (both task-based), 5 from DBOX (games collected in a Wizard-of-Oz fashion) and 4 from Switchboard (open-domain human-human conversation). Overall there are 1,596 DAs. The corpus currently represents the only publicly available resource manually annotated using the ISO standard.

\textbf{Common Alexa Prize Conversations (CAPC)}: The CAPC corpus \cite{ram2017} is a dataset of 3,764 anonymised individual user turns pooled from different users interacting with all socialbots participating in the Alexa Prize. We have extracted a balanced subset of 458 turns and have annotated it with DAs from our adapted version of the ISO standard by 3 annotators, with an inter-annotator agreement of $\kappa = 0.82$. CAPC exemplifies frequent user interaction data not biased by the interaction with one socialbot in particular. Another advantage of this dataset is that it is balanced across different DA categories. One drawback is that no interaction context (previous DA) is available for the individual turns.

\textbf{Socialbot Logs (S-Logs)}: S-Logs is a dataset of 13 open-domain conversations that different native American English speakers had with one of the socialbots of the Alexa Prize Challenge 2017. 
Overall this dataset contains 310 machine DAs and 165 user DAs. Two annotators tagged this dataset with DAs from our adapted version of ISO 24617-2, with an inter-annotator agreement of $\kappa = 0.81$. While we have annotated both machine and user turns, we test only on the latter and exploit machine turns as features for our classification experiments. 

\section{Methodology}
\label{sec:methodology}
\subsection{Preprocessing}
Before mapping the DA schemes of the corpora to the ISO subset scheme, a series of preprocessing steps have been performed to obtain a uniform training resource with the same surface text features as the testing corpora, since in S-Logs data, user input is lowercased and the punctuation is limited to apostrophes. More specifically, the text has been lowercased (including any named entity appearing in the original transcription and excluding the `I' pronoun), punctuation has been removed (except for the apostrophe character in contracted expressions like ``let's'' and ``can't'') and any special characters have been deleted from the utterances. Moreover, any information regarding prosody has been removed, since this feature is not available in our test sets. 

For experiments on the SWDA DAMSL corpus we recreated the same setting described in \cite{stolcke2006dialogue}, using the same train and test set and preprocessing the corpus in the same way following the WS97 manual annotator guidelines \cite{switchboardTags}.

\subsection{Dialogue Act scheme and mapping}
\begin{figure}
\centering
\scalebox{0.65}{
\begin{tikzpicture}[auto]
\Tree [.\textbf{General~(Task)}  
           [.\textbf{Information-Transfer~Functions}
             [.\textbf{Info-Seeking}
               [.Question
                 [.\textit{Prop~Q} \textcolor{gray}{Check~Q} ]
                 [.\textit{Set~Q} ]
                 [.\textit{Choice~Q} ]
               ]
             ]
             [.\textbf{Info-Providing}
               [.Inform
                 [.\textcolor{gray}{Answer}
                   [.\textcolor{gray}{Confirm} ]
                   [.\textcolor{gray}{Disconfirm} ]
                 ]
                 [.\textcolor{gray}{Agr.} ]
                 [.\textcolor{gray}{Disagr.} \textcolor{gray}{Correction} ]
               ]
             ]
           ]
           [.\textbf{Action-Discussion~Functions}
             [.\textbf{\textit{Commissives}} 
               [.\textcolor{gray}{Offer} 
                 [.\textcolor{gray}{Promise} \textcolor{gray}{A.Request~(A/D)} ]
               ]
               [.\textcolor{gray}{A.Suggest~(A/D)} ]
             ]
             [.\textbf{\textit{Directives}} 
               [.\textcolor{gray}{Suggest} ]
               [.\textcolor{gray}{Request}  
                 [.\textcolor{gray}{Instruct} \textcolor{gray}{A.Offer~(A/D)} ]
               ]
             ]  
           ] 
       ]

\end{tikzpicture}}
\label{fig:tree}
\caption{Communicative functions from ISO 24617-2 General purpose (Task) dimension. The nodes of the taxonomy that are not considered are grayed out.}
\end{figure}
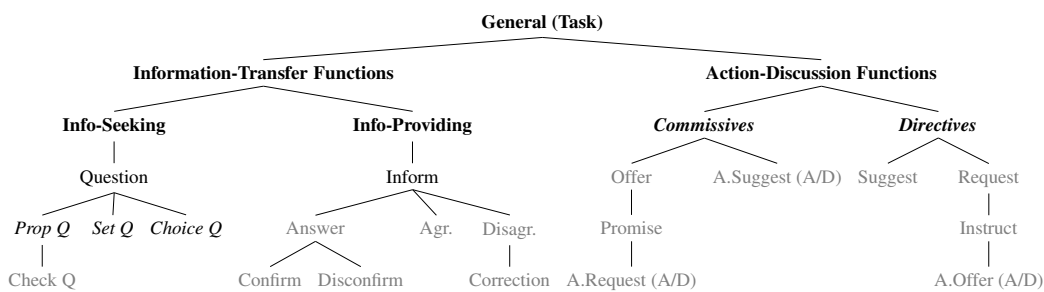

\begin{center}
\begin{table}
\begin{tabularx}{\columnwidth}{|X|X|}
\hline
\textbf{S-scheme } & \textbf{ISO 24617-2} \\
\hline\hline
\multicolumn{2}{|c|}{\textbf{Social Obligation Management}} \\
\hline
\textit{Salutation} & Greeting, Goodbye, Self-Intro\\
\textit{Apology}    & Apology, Accept Apology\\
\textit{Thanking}   & Thanking, Accept Thanking\\
\hline\hline
\multicolumn{2}{|c|}{\textbf{Feedback}} \\
\hline
\textit{Feedback}   & Auto-Feedback (all), Allo-Feedback (all)\\
\hline
\end{tabularx}
\caption{Our scheme (S-scheme) compared to the corresponding ISO 24617-2 scheme for the SOM and Feedback dimensions}
\label{table:somfb}
\end{table}
\end{center}

The Socialbot scheme (S-scheme), the DA scheme used during the classification experiments, is a subset of the official ISO standard. Only three dimensions out of the official eight defined in the standard are considered (\textit{Task}, \textit{Social Obligation Management} and \textit{Feedback}), and some of the communicative functions are generalized with an higher level of the tree.

Figure \ref{fig:tree} shows the labeled subset of the ISO standard taxonomy for the General-purpose functions (i.e. functions independent from any given dimensions), while table \ref{table:somfb} shows the correspondence for dimension-specific functions. The main difference between the DA scheme labeled in this work and the complete ISO taxonomy is the lack of further specification for the \textit{Inform}, \textit{Commissive} and \textit{Directive} tags. This is due to the fact that most of the DA schemes used when building the training set do not provide contextual information detailed enough to label these tags accurately. Moreover, there is confusion and discrepancies about when these contextual DA should be used, even in the official ISO guidelines. Indeed, in \cite{fang2012annotation}, which provides the official mapping from Switchboard to the ISO standard, it is reported that some contextual DA tags (for example \textit{other\_answer}) do not have a direct mapping to the standard. This becomes even more problematic considering that among the training resources there are corpora like AMI, MapTask or VerbMobil, which label answers as Informs, which would make training data for this class extremely noisy. A similar argument can be raised on the lower leaves of the \textit{Directive} and \textit{Commissive} nodes, some of which are not labeled even in the very detailed SWDA taxonomy. Mapping of the available corpora to this scheme was done according to the available documentation in literature. 

For the Switchboard corpus, a detailed mapping is provided in \cite{fang2012annotation}, which was followed exactly for the supported dimensions/communicative functions. For MapTask and AMI, there is already research highlighting similarities and differences between their schemes and the ISO standard one \cite{petukhova2014interoperability}. These results do not provide an exact mapping between the two schemes, which in some cases is impossible: for example the AMI \textit{Elicit-inform} tag is the equivalent of ISO 24617-2's \textit{Question}, but will not map to any specific question type (\textit{SetQ}, \textit{PropQ}, \textit{ChoiceQ}, etc.). Utterances whose tags cannot be directly mapped to the ISO scheme were dropped and do not appear in the training set. 

Since there is no available literature on mapping the VerbMobil 2 and BT Oasis corpora to the ISO standard, a specific mapping was designed from scratch by drawing inspiration from the approaches available on other corpora. 


\begin{table}
\begin{small}
\begin{tabularx}{\textwidth}{|l|R|R|R|R|R|R|R|R|}
\hline
\textbf{DA} 
& \multicolumn{1}{c|}{\textbf{SWDA}}
& \multicolumn{1}{c|}{\textbf{MapTask}} 
& \multicolumn{1}{c|}{\textbf{VerbMobil}} 
& \multicolumn{1}{c|}{\textbf{Oasis BT}} 
& \multicolumn{1}{c|}{\textbf{AMI}} 
& \multicolumn{1}{c|}{\textbf{DB}} 
& \multicolumn{1}{c|}{\textbf{CAPC}} 
& \multicolumn{1}{c|}{\textbf{S-Logs}} \\
\hline\hline
\multicolumn{9}{|c|}{\textbf{Semantic Dimensions}} \\
\hline
\textit{General (Task)}  &     83,652 & 15,054     &     5,330 &    2587 &  1,523 &  1035 &  442 &   142 \\
\textit{Social OM}   &      2,866 & 0 & 384 &   588 & 10,039 & 21 &  16 &  7 \\
\textit{Feedback}          & 39,866 & 5,070 & 2,768     & 1,172 & 31,985 & 407 & 0 &  16 \\
\hline
\textbf{Total$^*$}           & 126,384 & 20,508 & 8,482 & 2,381 & 43,547 & 855 & 329 & 109 \\
\hline\hline
\textbf{\% of Corpus} & 79\% & 100\% & 72\% & 58\% & 74\% & 100\% & 100\% & 100\% \\
\hline\hline

\multicolumn{9}{|c|}{\textbf{General (Task) Dimension}} \\
\hline
\textit{Commissives}  &     63 & -     &     7 &    25 &  1,523 &  57 &  20 &   1 \\
\textit{Directives}   &      7 & 4,075 & 2,911 &   181 & 10,039 & 131 &  93 &  32 \\
\textit{Inform}          & 75,667 & 4,860 & -     & 1,648 & 33,403 & 652 & 105 &  91 \\
\textit{Prop. Question}  &  1,986 &   583 & -     &   492 & -      &  61 &  68 &  12 \\    
\textit{Set Question}    &  5,506 & 1,692 & -     &   241 & -      & 134 & 149 &   6 \\
\textit{Choice Question} &    423 & -     & -     & -     & -      &   8 &   7 &  -  \\
\hline
\textbf{Total$^*$}           & 83,652 & 11,210 & 2,918 & 2,587 & 44,965 & 1,035 & 442 & 142 \\
\hline\hline
\textbf{\% of Corpus} & 57\% & 30\% & 32\% & 35\% & 34\% & N/A & N/A & N/A \\
\hline\hline

\multicolumn{8}{|c|}{\textbf{Social Obligations Management}} \\
\hline
\textit{Salutation} & 2,711 & - & 340 & 231 & -     & 13 &  6 & 2 \\
\textit{Apology}    &    75 & - &   - &  44 & -     &  6 &  3 & 4 \\ 
\textit{Thanking}   &    80 & - &  44 & 193 & -     &  2 &  7 & 1 \\ 
\hline
\textbf{Total$^*$}      & 2,866 & 0 & 384 & 468 & 2,201 & 21 & 16 & 7 \\
\textbf{\% of Corpus} & 2\% & 0\% & 2\% & 8\% & 0\% & N/A & N/A & N/A \\
\hline\hline

\multicolumn{9}{|c|}{\textbf{Feedback}} \\
\hline
\textbf{Total} & 39,886 & 5,070 & 2,768 & 1,172 & 31,985 & 407& -    &  16 \\
\textbf{\% of Corpus} & 79\% & 100\% & 72\% & 58\% & 74\% & N/A & N/A & N/A \\
\hline
\end{tabularx}
\end{small}
\caption{Dialogue Act category counts across the considered corpora for different levels of the taxonomy. \textit{Percentages of corpora} indicate the percentage of data available for the particular level in the corpus. $*$ It is frequently the case that DA tags do not map to any leaf-node, e.g. VerbMobil for Task dimension and AMI for Social Obligations Management. }\label{tbl:dadistrib}
\end{table}

Table \ref{tbl:dadistrib} presents counts of the DAs after mapping to our scheme, across all training and testing corpora. As mentioned in the paper, the corpora present quite imbalanced distributions of DA categories.

\section{Experiments and Results}
Since, to the best of our knowledge, there is no established state of the art on DialogBank -- the only corpus manually annotated following the ISO 24617-2 scheme -- we first establish the tagging methodology on the SWDA corpus using the DAMSL 42 tag set and compare it to the state of the art \cite{ji2016latent}. Then, the feature set and the parameters of the best performing models are used for the training of the DA tagger on the aggregate dataset, considering some of the semantic dimensions and the communicative functions of the ISO 24617-2 . The models are then evaluated on the DialogBank and open-domain human-machine data from Amazon Alexa Prize Challenge. McNemar's test \cite{McNemar1947} for statistical significance has been used to analyze whether introduced features give a significant contribution to the overall performance.

\subsection{Experiments on SWDA}
Prior to training the classification models, the SWDA \cite{switchboardTags} utterances are preprocessed following \cite{stolcke2006dialogue}. The dataset is split into training (1,115 dialogues) and test set (19 dialogues) following the same paper, and the remaining 21 dialogues are used as development set to tune the $C$ parameter of Support Vector Machines (SVM) \cite{SVM}. For the experiments, we used the SVM implementation of scikit-learn \cite{scikit-learn} with linear kernel (i.e. its \textit{liblinear} \cite{liblinear} wrapper).

The results of the experiments on SWDA are presented in Table \ref{table:SWDAresults}. The performances are on the SWDA test set with the SVM $C$ parameter set to 0.1, with respect to the best results on the development set. It is worth mentioning that tuning the $C$ parameter boosts the performance on the development set by 2 points.\footnote{From 73 ($C=1.0$) to 75 ($C=0.1$) for the model trained on unigrams, bigrams and previous DA-tag.} For comparison, the table also includes majority baseline, the results from \cite{stolcke2006dialogue}, the SVM results from \cite{QuarteroniR10}, and the state-of-the-art results from \cite{ji2016latent} that were achieved using deep learning methods. 

\begin{table}
\centering
\begin{tabularx}{0.49\columnwidth}{|l|Rc|}
\hline
\textbf{Features}  & \multicolumn{2}{c|}{\textbf{Acc.}} \\
\hline\hline
BL: Majority & 31.5 & \\ 
\hline
HMM \cite{stolcke2006dialogue}  & 71.0 & \\
\hline
SVM \cite{QuarteroniR10}        & 72.4 & \\
DrLM (LSTM) \cite{ji2016latent} & 77.0 & \\
\hline
1-grams      & 71.2 & \\ 
1-2-grams    & 71.7 & \\ 
1-2-3-grams  & 71.4 & \\ 
\hline
\end{tabularx} \begin{tabularx}{0.49\columnwidth}{|l|Rc|}
\hline
\textbf{Features}  & \multicolumn{2}{c|}{\textbf{Acc.}} \\
\hline\hline
~ & ~ & \\
\hline
1-2-grams + PREV & 74.6 & * \\ 
\hline
1-2-grams + PREV + POS    & 74.6 & \\ 
1-2-grams + PREV + I-POS  & 76.2 & * \\ 
\hline
1-2-grams + PREV + I-POS + DEP   & 76.0 & \\ 
1-2-grams + PREV + I-POS + I-DEP & 76.1 & \\ 
1-2-grams + PREV + I-POS + WE & \textbf{76.7} & * \\ 
\hline
\end{tabularx}
\caption{Classification accuracy of the different feature combinations on the SWDA test set. The best results are highlighted in bold. The results that are significantly better are marked with *.}
\label{table:SWDAresults}
\end{table}

Following the previous studies on SWDA \cite{stolcke2006dialogue,QuarteroniR10}, we experiment with n-grams (unigrams, bigrams, and trigrams) and previous DA tag features. We do not consider the unit length feature from \cite{QuarteroniR10}, since classification instances in the SWDA scheme and ISO 24617-2 are different (slash unit vs. functional unit). The results are reported in Table \ref{table:SWDAresults}; since the results reported were obtained with SVM $C$ parameter set to 0.1, they are higher than the ones reported in \cite{QuarteroniR10}: e.g. for 1-2-grams 70.0 vs. 71.7.

The first observation is that the addition of the previous DA significantly improves the performance. Addition of part-of-speech tags does not yield any improvement; however, when POS-tags are indexed with their positions in an utterance, accuracy is significantly improved and rises to 76.2. Addition of dependency relations (both with and without indexing with their position) does not improve the performance. Addition of the averaged pre-trained word-embedding vectors (from Google News) to the model with indexed POS-tags, however, rises the accuracy to 76.7. The model with word embeddings comes 0.3 short of the state-of-the-art results reported in \cite{ji2016latent}.

\subsection{Experiments on Aggregate ISO-standard Data}
The methodology established on SWDA is applied to training the ISO 24617-2 subset models using the aggregate data set. Since in ISO 24617-2 annotation scheme DAs consist of semantic dimensions and communicative functions, the utterances are first classified into the considered semantic dimensions -- general, social obligations management (SOM), and feedback. Then, we experiment with the Task dimension, reporting the results without error propagation from the previous step, in order to give the reader a clearer understanding of the current classification capabilities when restricting interactions with the system to general communicative functions.

\subsubsection{Semantic Dimension Classification}
The results of the binary dimension classification models on the test sets -- DialogBank (DB), CAPC, and S-Logs -- are reported in Table \ref{table:dimensionAccuracy}. The CAPC corpus consists of isolated utterances; consequently, the \textit{Feedback} dimension is not present. On DB and S-Logs, on the other hand, the \textit{Feedback} dimension yields the lowest accuracy in comparison to General and \textit{SOM} dimension communicative functions. Low performances on the \textit{Feedback} dimension could be explained by the fact that the training data mostly contains \textit{Allo-feedback} and lacks \textit{Auto-feedback} and \textit{Feedback elicitations}, which are present in DB.

\begin{table}
\centering
\begin{tabularx}{\columnwidth}{|l|R|R|R|}
\hline
\textbf{Dimension}
& \multicolumn{1}{c|}{\textbf{DB}}
& \multicolumn{1}{c|}{\textbf{CAPC}}
& \multicolumn{1}{c|}{\textbf{S-Logs}} \\
\hline
\textit{General}  & 73.3 & 83.0 & 80.2 \\
\textit{SOM}      & 78.1 & 90.7 & 86.6 \\
\textit{Feedback} & 56.3 & --   & 71.3 \\
\hline
\textit{Overall}  & 68.4 & 83.3 & 79.4 \\
\hline
\end{tabularx}
\caption{Classification accuracies of the binary semantic dimension models: General, SOM and Feedback. The CAPC corpus does not contain Feedbacks, therefore results for this dimension are not reported.}
\label{table:dimensionAccuracy}
\end{table}

\subsubsection{Communicative Function Classification}

\begin{table}
\centering
\begin{tabularx}{0.49\columnwidth}{|l|R|R|R|}
\hline
\textbf{Features}
& \multicolumn{1}{c|}{\textbf{DB}} 
& \multicolumn{1}{c|}{\textbf{CAPC}} 
& \multicolumn{1}{c|}{\textbf{S-Logs}}\\
\hline\hline
BL: Majority & 53.4\textcolor{white}{*} & 22.9\textcolor{white}{*} & 63.4\textcolor{white}{*} \\
\hline\hline
1-2-grams    & 64.2\textcolor{white}{*} & 71.2\textcolor{white}{*} & 78.7\textcolor{white}{*} \\
+ PREV  & 64.3\textcolor{white}{*} & 70.7\textcolor{white}{*} & 81.6\textcolor{black}{*} \\
+ I-POS & 65.8\textcolor{black}{*} & 73.8\textcolor{black}{*} & 82.2\textcolor{white}{*} \\ 
\hline
\end{tabularx} \begin{tabularx}{0.49\columnwidth}{|l|R|R|R|}
\hline
\textbf{Features}
& \multicolumn{1}{c|}{\textbf{DB}} 
& \multicolumn{1}{c|}{\textbf{CAPC}} 
& \multicolumn{1}{c|}{\textbf{S-Logs}}\\
\hline\hline
~ &&&\\
\hline\hline
+ I-DEP & 67.1\textcolor{black}{*} & 74.3\textcolor{white}{*} & 82.3\textcolor{white}{*} \\
+ WE & 65.2\textcolor{white}{*} & 74.8\textcolor{black}{*} &82.0\textcolor{white}{*} \\
+ I-DEP + WE & 66.6\textcolor{white}{*} & 75.1\textcolor{white}{*} & 81.8\textcolor{white}{*} \\

\hline
\end{tabularx}
\caption{Accuracies of the feature combinations on the general-purpose communicative functions on the test sets. The best results are marked in bold, and statistically significant differences with *.}
\label{table:SVMresults}
\end{table}



The utterances are further classified into communicative functions of the General (Task) dimension, using the methodology established on SWDA, i.e. the same hyper-parameter settings ($C=0.1$) and features. However, since models with dependency relations do not yield statistically significant differences, they are also considered. The results of the models on the test sets are reported in Table \ref{table:SVMresults}.
The behavior of the models trained with various feature combinations is in-line with the SWDA experiments: the addition of the previous DA tags and part-of-speech tags indexed with their positions in a sentence improves the performance. Different from the SWDA, the addition of the indexed dependency relations improves the performance on the test sets. In the case of DialogBank and CAPC, their contribution is statistically significant.
Additionally, unlike for SWDA, the addition of word embeddings with and without index dependency relation (I-DEP) does not produce significant improvements for all but CAPC. Consequently, the model trained on 1-2-grams, previous DA tags, indexed POS-tags and dependency relations is chosen for the ablation study.

\subsubsection{Corpora Combinations}
The aggregation of all the corpora mapped to our subset of ISO 24617-2 is not necessarily the best one, as the distributions of DA categories varies from corpus to corpus. Consequently, we also present results on the test sets for the models trained solely on SWDA and AMI; as well as perform an ablation experiment removing one corpus at a time. The best performing model from the previous subsection (1-2-grams, previous DA-tag, indexed POS-tags, and indexed dependency relations) is used for the study. The results of these experiments are reported in Table \ref{table:dataset_contribution}. 

While the best results for Dialog Bank are achieved considering all the corpora, for CAPC the best results are achieved by removing MapTask. For S-Logs, on the other hand, the best performing corpora combination is all except VerbMobil. However, the performance differences from the models trained on all corpora are not statistically significant. Training DA taggers solely on SWDA and AMI -- the largest and the most diverse corpora -- yields performances inferior to the combination of all the corpora. From the table, we can also observe that these two corpora -- SWDA and AMI -- contribute most to the performance, as removing them affects the performance the most. On the other hand, removing the smaller datasets -- BT Oasis, MapTask, and VerbMobil -- affects the performance less.

\begin{table}
\centering
\begin{tabularx}{0.49\columnwidth}{|l|R|R|R|}
\hline
\textbf{Dataset}
& \multicolumn{1}{c|}{\textbf{DB}} 
& \multicolumn{1}{c|}{\textbf{CAPC}} 
& \multicolumn{1}{c|}{\textbf{S-Logs}}\\
\hline\hline
ALL           &\textbf{67.1} & 74.3 & 82.3 \\
~ &&&\\
SWDA only     & 57.9 & 71.3 & 53.5 \\ 
AMI only      & 53.2 & 39.8 & 61.6 \\
~ &&&\\
\hline
\end{tabularx} \begin{tabularx}{0.49\columnwidth}{|l|R|R|R|} 
\hline
\textbf{Dataset}
& \multicolumn{1}{c|}{\textbf{DB}} 
& \multicolumn{1}{c|}{\textbf{CAPC}} 
& \multicolumn{1}{c|}{\textbf{S-Logs}}\\
\hline\hline
- AMI         & 59.7 & 73.7 & 71.3 \\
- SWDA 		  & 60.2 & 68.3 & 77.5 \\ 
- Oasis BT    & 66.1 & 74.2 & 81.8 \\ 
- MapTask     & 66.8 & \textbf{74.6} & 80.5 \\
- VerbMobil   & 66.5 & 74.0 & \textbf{82.6} \\ 
\hline
\end{tabularx}
\caption{Accuracies of the corpora combinations on the test sets -- Dialog Bank (DB), CAPC, and S-Logs.}
\label{table:dataset_contribution}
\end{table}

\section{Conclusions}

We have presented an effective methodology for corpora aggregation for domain-independent Dialogue Act Tagging on a subset of the ISO 24617-2 annotation. We have also reported an accurate evaluation of our approach on both in-domain and out-of-domain datasets, proving that the described DA tagging technique is indeed independent from the underlying scheme and task of the annotated corpora. Finally, the machine learning technique used for DA tagging was tested on a popular DA tagging task (the Switchboard corpus), obtaining very close to state-of-the-art results. 

This work represents one of the first attempts to use an ISO compliant DA scheme for a real-life application, as well as one of the first structured approaches for evaluation of dialogue resources annotated with this taxonomy. 

Research on available training resources is one of the first things to look forward to, since the current data proved to be effective, but also presented numerous drawbacks (lack of adequate coverage for the Feedback dimension, imbalanced DAs, lack of context-aware communicative functions). We plan to make our resource continue to grow in the future by adding and mapping additional corpora, such as the MRDA \cite{shriberg2004icsi} corpus. 

\section{Acknowledgments}
This project was partially funded by Amazon Alexa Prize 2017 research grant.

\bibliography{strings}

\begin{thebibliography}{}

\bibitem[\protect\citename{Alexandersson \bgroup et al.\egroup
  }1998]{verbmobil2}
Jan Alexandersson, Bianka Buschbeck-Wolf, Tsutomu Fujinami, Michael Kipp,
  Stephan Koch, Elisabeth Maier, Norbert Reithinger, Birte Schmitz, and Melanie
  Siegel.
\newblock 1998.
\newblock {\em Dialogue acts in Verbmobil 2}.
\newblock DFKI Saarbr{\"u}cken.

\bibitem[\protect\citename{Anderson \bgroup et al.\egroup
  }1991]{anderson1991hcrc}
Anne~H Anderson, Miles Bader, Ellen~Gurman Bard, Elizabeth Boyle, Gwyneth
  Doherty, Simon Garrod, Stephen Isard, Jacqueline Kowtko, Jan McAllister, Jim
  Miller, et~al.
\newblock 1991.
\newblock The hcrc map task corpus.
\newblock {\em Language and speech}, 34(4):351--366.

\bibitem[\protect\citename{Austin and Urmson}1962]{austin1962things}
John~Langshaw Austin and JO~Urmson.
\newblock 1962.
\newblock {\em How to Do Things with Words. The William James Lectures
  Delivered at Harvard University in 1955.[Edited by James O. Urmson.].}
\newblock Clarendon Press.

\bibitem[\protect\citename{Bowden \bgroup et al.\egroup
  }2017]{bowden2018slugbot}
Kevin~K Bowden, Jiaqi Wu, Shereen Oraby, Amita Misra, and Marilyn Walker.
\newblock 2017.
\newblock Slugbot: An application of a novel and scalable open domain socialbot
  framework.
\newblock {\em Alexa Prize Proceedings}.

\bibitem[\protect\citename{Bunt \bgroup et al.\egroup }2010]{buntISOstandard}
Harry Bunt, Jan Alexandersson, Jean Carletta, Jae-Woong Choe, Alex~Chengyu
  Fang, Koiti Hasida, Kiyong Lee, Volha Petukhova, Andrei Popescu-Belis,
  Laurent Romary, Claudia Soria, and David Traum.
\newblock 2010.
\newblock Towards an iso standard for dialogue act annotation.
\newblock {\em Seventh conference on International Language Resources and
  Evaluation (LREC'10)}.

\bibitem[\protect\citename{Bunt \bgroup et al.\egroup }2012]{bunt2012iso}
Harry Bunt, Jan Alexandersson, Jae-Woong Choe, Alex~Chengyu Fang, Koiti Hasida,
  Volha Petukhova, Andrei Popescu-Belis, and David~R Traum.
\newblock 2012.
\newblock Iso 24617-2: A semantically-based standard for dialogue annotation.
\newblock In {\em LREC}, pages 430--437.

\bibitem[\protect\citename{Bunt \bgroup et al.\egroup
  }2016]{bunt2016dialogbank}
Harry Bunt, Volha Petukhova, Andrei Malchanau, Kars Wijnhoven, and Alex~Chengyu
  Fang.
\newblock 2016.
\newblock The dialogbank.
\newblock In {\em LREC}.

\bibitem[\protect\citename{Bunt}1999]{bunt1999dynamic}
Harry Bunt.
\newblock 1999.
\newblock Dynamic interpretation and dialogue theory.
\newblock {\em The structure of multimodal dialogue}, 2:1--8.

\bibitem[\protect\citename{Bunt}2009]{bunt2009dit++}
Harry Bunt.
\newblock 2009.
\newblock The dit++ taxonomy for functional dialogue markup.
\newblock In {\em AAMAS 2009 Workshop, Towards a Standard Markup Language for
  Embodied Dialogue Acts}, pages 13--24.

\bibitem[\protect\citename{Burger \bgroup et al.\egroup
  }2000]{burger2000verbmobil}
Susanne Burger, Karl Weilhammer, Florian Schiel, and Hans~G Tillmann.
\newblock 2000.
\newblock Verbmobil data collection and annotation.
\newblock In {\em Verbmobil: Foundations of speech-to-speech translation},
  pages 537--549. Springer.

\bibitem[\protect\citename{Carletta}2006]{amicorpus}
Jean Carletta.
\newblock 2006.
\newblock Announcing the ami meeting corpus.
\newblock {\em The ELRA Newsletter 11(1), January-March, p. 3-5.}

\bibitem[\protect\citename{Cervone \bgroup et al.\egroup }2017]{cervoneroving}
Alessandra Cervone, Giuliano Tortoreto, Stefano Mezza, Enrico Gambi, and
  Giuseppe Riccardi.
\newblock 2017.
\newblock Roving mind: a balancing act between open--domain and engaging
  dialogue systems.
\newblock In {\em Alexa Prize Proceedings}.

\bibitem[\protect\citename{Chowdhury \bgroup et al.\egroup
  }2016]{chowdhury2016transfer}
Shammur~Absar Chowdhury, Evgeny~A Stepanov, and Giuseppe Riccardi.
\newblock 2016.
\newblock Transfer of corpus-specific dialogue act annotation to iso standard:
  Is it worth it?
\newblock In {\em LREC}.

\bibitem[\protect\citename{Core and Allen}1997]{DAMSL}
Mark~G. Core and James~F. Allen.
\newblock 1997.
\newblock Coding dialogs with the damsl annotation scheme.
\newblock In {\em Proceedings of AAAI Fall Symposium on Communicative Action in
  Humans and Machines}.

\bibitem[\protect\citename{Fan \bgroup et al.\egroup }2008]{liblinear}
Rong-En Fan, Kai-Wei Chang, Cho-Jui Hsieh, Xiang-Rui Wang, and Chih-Jen Lin.
\newblock 2008.
\newblock Liblinear: A library for large linear classification.
\newblock {\em J. Mach. Learn. Res.}, 9:1871--1874, June.

\bibitem[\protect\citename{Fang \bgroup et al.\egroup
  }2012]{fang2012annotation}
Alex~C. Fang, Jing Cao, Harry Bunt, and Xiaoyue Liu.
\newblock 2012.
\newblock The annotation of the {Switchboard Corpus} with the new {ISO}
  standard for dialogue act analysis.
\newblock In {\em Workshop on Interoperable Semantic Annotation}.

\bibitem[\protect\citename{Fang \bgroup et al.\egroup }2017]{soundingboard2017}
Hao Fang, Hao Cheng, Elizabeth Clark, Ariel Holtzman, Maarten Sap, Mary
  Ostendorf, Yejin Choi, and Noah~A. Smith.
\newblock 2017.
\newblock Sounding board -- university of washington's alexa prize submission.
\newblock In {\em Alexa Prize Proceedings}.

\bibitem[\protect\citename{Godfrey \bgroup et al.\egroup
  }1992]{godfrey1992switchboard}
John~J Godfrey, Edward~C Holliman, and Jane McDaniel.
\newblock 1992.
\newblock Switchboard: Telephone speech corpus for research and development.
\newblock In {\em Acoustics, Speech, and Signal Processing, 1992. ICASSP-92.,
  1992 IEEE International Conference on}, volume~1, pages 517--520. IEEE.

\bibitem[\protect\citename{Gupta \bgroup et al.\egroup }2006]{gupta2006t}
Narendra Gupta, Gokhan Tur, Dilek Hakkani-Tur, Srinivas Bangalore, Giuseppe
  Riccardi, and Mazin Gilbert.
\newblock 2006.
\newblock The at\&t spoken language understanding system.
\newblock {\em IEEE Transactions on Audio, Speech, and Language Processing},
  14(1):213--222.

\bibitem[\protect\citename{Henderson \bgroup et al.\egroup
  }2014]{henderson2014second}
Matthew Henderson, Blaise Thomson, and Jason~D Williams.
\newblock 2014.
\newblock The second dialog state tracking challenge.
\newblock In {\em SIGDIAL Conference}, pages 263--272.

\bibitem[\protect\citename{Ji \bgroup et al.\egroup }2016]{ji2016latent}
Yangfeng Ji, Gholamreza Haffari, and Jacob Eisenstein.
\newblock 2016.
\newblock A latent variable recurrent neural network for discourse relation
  language models.
\newblock In {\em Proceedings of NAACL-HLT}, pages 332--342.

\bibitem[\protect\citename{Jurafsky}1997]{switchboardTags}
Dan Jurafsky.
\newblock 1997.
\newblock Switchboard swbd-damsl shallow-discourse-function.
\newblock {\em Annotation, Technical Report, 97-02, University of Colorado, CO,
  USA.}

\bibitem[\protect\citename{Kumar \bgroup et al.\egroup
  }2017]{kumar2017dialogue}
Harshit Kumar, Arvind Agarwal, Riddhiman Dasgupta, Sachindra Joshi, and Arun
  Kumar.
\newblock 2017.
\newblock Dialogue act sequence labeling using hierarchical encoder with crf.
\newblock {\em arXiv preprint arXiv:1709.04250}.

\bibitem[\protect\citename{Leech and Weisser}2003]{leech2003generic}
Geoffrey Leech and Martin Weisser.
\newblock 2003.
\newblock Generic speech act annotation for task-oriented dialogues.
\newblock In {\em Procs. of the 2003 Corpus Linguistics Conference, pp.
  441Y446. Centre for Computer Corpus Research on Language Technical Papers,
  Lancaster University}.

\bibitem[\protect\citename{McNemar}1947]{McNemar1947}
Quinn McNemar.
\newblock 1947.
\newblock Note on the sampling error of the difference between correlated
  proportions or percentages.
\newblock {\em Psychometrika}, 12(2):153--157, Jun.

\bibitem[\protect\citename{Pedregosa \bgroup et al.\egroup }2011]{scikit-learn}
F.~Pedregosa, G.~Varoquaux, A.~Gramfort, V.~Michel, B.~Thirion, O.~Grisel,
  M.~Blondel, P.~Prettenhofer, R.~Weiss, V.~Dubourg, J.~Vanderplas, A.~Passos,
  D.~Cournapeau, M.~Brucher, M.~Perrot, and E.~Duchesnay.
\newblock 2011.
\newblock Scikit-learn: Machine learning in {P}ython.
\newblock {\em Journal of Machine Learning Research}.

\bibitem[\protect\citename{Petukhova \bgroup et al.\egroup
  }2014a]{petukhova2014dbox}
Volha Petukhova, Martin Gropp, Dietrich Klakow, Anna Schmidt, Gregor Eigner,
  Mario Topf, Stefan Srb, Petr Motlicek, Blaise Potard, John Dines, et~al.
\newblock 2014a.
\newblock The dbox corpus collection of spoken human-human and human-machine
  dialogues.
\newblock In {\em Proceedings of the Ninth International Conference on Language
  Resources and Evaluation (LREC" 14)}, number EPFL-CONF-201766. European
  Language Resources Association (ELRA).

\bibitem[\protect\citename{Petukhova \bgroup et al.\egroup
  }2014b]{petukhova2014interoperability}
Volha Petukhova, Andrei Malchanau, and Harry Bunt.
\newblock 2014b.
\newblock Interoperability of dialogue corpora through {ISO} 24617-2-based
  querying.
\newblock In {\em LREC}.

\bibitem[\protect\citename{Price}1990]{price1990evaluation}
Patti~J Price.
\newblock 1990.
\newblock Evaluation of spoken language systems: The atis domain.
\newblock In {\em Speech and Natural Language: Proceedings of a Workshop Held
  at Hidden Valley, Pennsylvania, June 24-27, 1990}.

\bibitem[\protect\citename{Quarteroni and Riccardi}2010]{QuarteroniR10}
Silvia Quarteroni and Giuseppe Riccardi.
\newblock 2010.
\newblock Classifying dialog acts in human-human and human-machine spoken
  conversations.
\newblock In {\em {INTERSPEECH} 2010, 11th Annual Conference of the
  International Speech Communication Association}, pages 2514--2517.

\bibitem[\protect\citename{Quarteroni \bgroup et al.\egroup
  }2011]{quarteroni2011simultaneous}
Silvia Quarteroni, Alexei~V Ivanov, and Giuseppe Riccardi.
\newblock 2011.
\newblock Simultaneous dialog act segmentation and classification from
  human-human spoken conversations.
\newblock In {\em Acoustics, Speech and Signal Processing (ICASSP), 2011 IEEE
  International Conference on}, pages 5596--5599. IEEE.

\bibitem[\protect\citename{Ram \bgroup et al.\egroup }2017]{ram2017}
Ashwin Ram, Rohit Prasad, Chandra Khatri, Anu Venkatesh, Raefer Gabriel, Qing
  Liu, Jeff Nunn, Behnam Hedayatnia, Ming Cheng, Ashish Nagar, Eric King, Kate
  Bland, Amanda Wartick, Yi~Pan, Han Song, Sk~Jayadevan, Gene Hwang, and Art
  Pettigrue.
\newblock 2017.
\newblock Conversational ai: The science behind the alexa prize.
\newblock In {\em Alexa Prize Proceedings}.

\bibitem[\protect\citename{Shriberg \bgroup et al.\egroup
  }2004]{shriberg2004icsi}
Elizabeth Shriberg, Raj Dhillon, Sonali Bhagat, Jeremy Ang, and Hannah Carvey.
\newblock 2004.
\newblock The icsi meeting recorder dialog act (mrda) corpus.
\newblock Technical report, INTERNATIONAL COMPUTER SCIENCE INST BERKELEY CA.

\bibitem[\protect\citename{Stolcke \bgroup et al.\egroup
  }2000]{stolcke2006dialogue}
Andreas Stolcke, Klaus Ries, Noah Coccaro, Elizabeth Shriberg, Rebecca Bates,
  Daniel Jurafsky, Paul Taylor, Rachel Martin, Carol Van Ess-Dykema, and Marie
  Meteer.
\newblock 2000.
\newblock Dialogue act modeling for automatic tagging and recognition of
  conversational speech.
\newblock {\em Computational Linguistics}, 26(3).

\bibitem[\protect\citename{Traum}1996]{traum1996conversational}
David Traum.
\newblock 1996.
\newblock Conversational agency: The trains-93 dialogue manager.
\newblock In {\em In Susann LuperFoy, Anton Nijhholt, and Gert Veldhuijzen van
  Zanten, editors, Proceedings of Twente Workshop on Language Technology,
  TWLT-II}. Citeseer.

\bibitem[\protect\citename{Vapnik}1995]{SVM}
V.N. Vapnik.
\newblock 1995.
\newblock {\em The Nature of Statistical Learning Theory}.
\newblock Springer.

\bibitem[\protect\citename{Xu and Sarikaya}2013]{xu2013convolutional}
Puyang Xu and Ruhi Sarikaya.
\newblock 2013.
\newblock Convolutional neural network based triangular crf for joint intent
  detection and slot filling.
\newblock In {\em Automatic Speech Recognition and Understanding (ASRU), 2013
  IEEE Workshop on}, pages 78--83. IEEE.

\bibitem[\protect\citename{Yang \bgroup et al.\egroup }2017]{yang2016end}
Xuesong Yang, Yun-Nung Chen, Dilek Hakkani-T{\"u}r, Paul Crook, Xiujun Li,
  Jianfeng Gao, and Li~Deng.
\newblock 2017.
\newblock End-to-end joint learning of natural language understanding and
  dialogue manager.
\newblock In {\em Acoustics, Speech and Signal Processing (ICASSP), 2017 IEEE
  International Conference on}, pages 5690--5694. IEEE.

\end{thebibliography}
\bibliographystyle{acl}

\newpage
\appendix
\section{Appendix A. Dialogue Acts mappings}
\label{sec:supplemental}

\begin{table*}[!h]
\centering
\begin{small}
\begin{tabularx}{\textwidth}{|l|X|X|X|X|X|}
\hline
\textbf{ISO}
& \textbf{SWDA} 
& \textbf{MapTask} 
& \textbf{VerbMobil} 
& \textbf{Oasis BT} 
& \textbf{AMI}\\
\hline\hline
\multicolumn{6}{|c|}{\textbf{Task}}\\
\hline\hline
\textit{Inform} & Statement-non-opinion, Statement-opinion, Rhetorical-question, Statement expanding y/n answer, Hedge  & explain, clarify & -- & Inform & Inform\\
\hline
\textit{ChoiceQ} & Or-question \newline Or-clause & -- & -- & -- & -- \\
\hline
\textit{SetQ} & Wh-question \newline Declarative\newline wh-question & query\_w & -- & q\_wh & -- \\
\hline
\textit{PropQ} & Yes-no-question, 
Backchannel\newline in question form, Tag-question, 
Declarative\newline Yes-no-question & query\_yn & -- & q\_yn & -- \\
\hline
\textit{Commissive} & Offer, Commit & - & Offer, Commit & Offer & Offer\\
\hline
\textit{Directive}  & Open-Option & Instruct & Request (all),\newline Suggest & Suggest, imp & Suggest,\newline Elicit-offer\\
\hline\hline
\multicolumn{6}{|c|}{\textbf{Social Obligation Management}}\\
\hline\hline
\textit{Thanking}  & Thanking, \newline You're-welcome & -- & -- & thank & -- \\
\hline
\textit{Apology} & Apology \newline Downplayer & -- & -- & pardon\newline regret & -- \\
\hline
\textit{Salutation} & Conventional- closing & -- & -- & bye, greet & -- \\
\hline\hline
\multicolumn{6}{|c|}{\textbf{Feedback}}\\
\hline\hline
\textit{Feedback} & Signal-not-understanding, \newline Acknowledge (backchannel), \newline Acknowledge answer, \newline Appreciation, \newline
Sympathy, \newline Summarize/ reformulate, \newline Repeat-phrase & Acknowledge & Feedback (all) & ackn & Backchannel\\
\hline
\end{tabularx}
\caption{Mapping for Dialogue Acts from each individual corpus to ISO DA-tags.}
\label{table:mapping}
\end{small}
\end{table*}



\end{document}